\documentclass[letterpaper]{article} 
\usepackage[preprint]{arxiv2026}

\usepackage[hyphens]{url}
\usepackage{graphicx} 
\usepackage{natbib} 
\usepackage{caption} 
\usepackage{booktabs}
\usepackage{amsmath,amssymb}
\usepackage{array}
\usepackage{xspace}

\makeatletter
\renewcommand{\@seccntformat}[1]{%
  \csname the#1\endcsname.\quad}
\makeatother

\newcommand{\method}{{Spectrum}\xspace}

\newcommand{\auc}{AUROC}
\usepackage[table]{xcolor}

\title{Learning How Much, Not Just What: Cross-Patient Burden Order for CT Vision-Language Pretraining}

\author{
Guoliang You\textsuperscript{1},
Haifan Gong\textsuperscript{2},
Xiaomeng Chu\textsuperscript{3}\corresponding
}

\affiliations{
\textsuperscript{1}University of Pennsylvania,
\textsuperscript{2}Harvard Medical School
\textsuperscript{3}Yale University
}

\begin{document}
\maketitle

\begin{abstract}

Volumetric CT vision–language pretraining learns 3D representations from scan–report pairs, but global and anatomy-aware objectives supervise only correspondence: they establish what is present and leave how much unconstrained. Nothing separates a mild from an extensive case of the same finding along a consistent direction, so the graded burden language in reports collapses into a present/absent signal. Longitudinal supervision would supply this order, but patient-matched CT pairs are scarce at scale; cross-sectional cohorts already encode weak burden cues across different patients. 
We introduce \method, an anatomy-conditioned framework that represents each study at whole-study and organ scopes. For each organ-mapped pathology, a rule-based scorer mines confidence-filtered lower-to-higher pairs of different patients, and Burden-Direction Alignment (BDA) aligns the pathology-conditioned image delta with the report delta at each scope, separating that direction from its reverse. Because the endpoints are different people, a target-conditioned aligner first makes them comparable, so the delta reflects burden rather than between-patient variation. 
BDA further separates the selected direction from its reverse, anchors it to the observed higher-burden endpoint, and enforces consistency across ordered triplets. Since every pair is drawn within a single pathology, BDA is designed to constrain intra-class structure that image–report contrast alone never touches.
\method attains 85.6 zero-shot AUROC on CT-RATE and 72.7 on external RAD-ChestCT, with consistent gains in linear probing and retrieval. Weak cross-patient order is thus a scalable complement to anatomy-aware correspondence, yielding burden-aware CT representations without longitudinal data.
\end{abstract}

\section{Introduction}
\label{sec:introduction}

Computed tomography is a cornerstone of clinical imaging, yet reading a volumetric study means reviewing hundreds of slices and integrating evidence across several anatomical systems~\citep{langlotz2024merlin,gong2024intensity,huang2025bcnet,wang2026costal,chen2023cancerunit}.
Volumetric CT vision–language pretraining (VLP) therefore offers a scalable route to transferable 3D representations, learning from the scan–report pairs that hospitals already produce~\citep{hamamci2024generalist,langlotz2024merlin,chen2023cancerunit}. 
What makes chest CT reports particularly valuable is not only that they name findings and their locations, but that they grade them: a nodule is recorded as smaller than 5 mm or as 66 × 55 mm, consolidation as mild or as extensive. This graded language costs nothing to collect, exists at the scale of entire cohorts. Nevertheless, current objectives compress it into a binary present/absent signal.

\begin{figure}[!t]
\centering
\includegraphics[width=\linewidth]{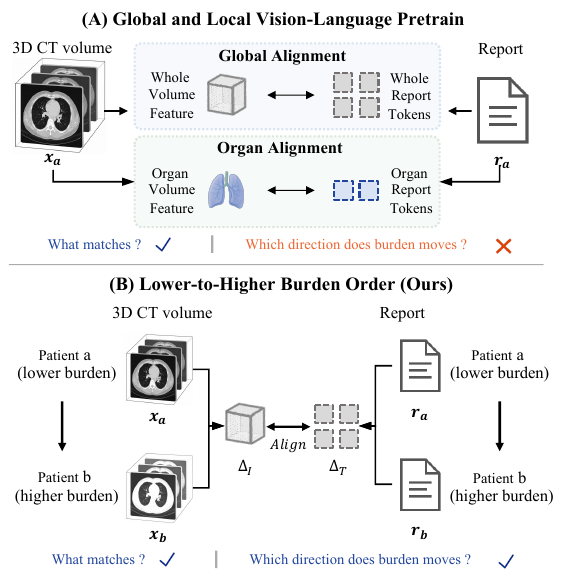}
\caption{
Correspondence versus burden order. (A) Global and organ objectives match a volume to its report at two anatomical scopes. (B) \method mines ordered pairs of different patients, $a$ (lower burden) and $b$ (higher), and aligns image delta $\Delta_I$ with report delta $\Delta_T$ at both scopes. Correspondence fixes \emph{what matches}; the order supplies \emph{how much}.
}
\label{fig:motivation}
\end{figure}

The dominant formulation aligns a whole volume with a whole report~\citep{hamamci2024generalist,langlotz2024merlin}, and recent methods extend this to local or anatomy-aware correspondence~\citep{shui2025fvlm}. 
These objectives answer one question well: which visual and textual observations belong together? But correspondence alone is directionless. A model may learn that lung evidence matches a lung snippet with no constraint that ``extensive'' disease lie beyond ``mild'' disease along a consistent pathology-burden direction.
The cost is a representation organized by category but not by degree: cases sharing a pathology are pulled toward a common region of feature space regardless of how much disease they contain, so anatomy localizes correspondence while burden order within that anatomy is left unconstrained.

Direct longitudinal supervision could impose such order, but repeat CT studies carrying comparable pathology evidence are difficult to assemble at the scale VLP requires. Cross-sectional cohorts, in contrast, are both large and already annotated in the relevant way: labels, negation, size mentions, burden descriptors, and organ-specific report fragments jointly indicate which of two patients carries more of a given pathology. This motivates our central question: \emph{can weak cross-patient order extend anatomy-aware correspondence toward burden-aware representations?}

Our key insight is to use anatomy as the common frame for both correspondence and order, so that each mined relation constrains burden direction within the same scopes in which correspondence is learned. Using two different patients as endpoints, however, raises a difficulty. The difference between their features reflects not only burden but also body habitus, acquisition protocol, and anatomical variation, none of which should define a burden direction. We therefore do not treat the raw feature difference as the supervision target: a target-conditioned aligner first makes the lower-burden representation comparable with its higher-burden counterpart, and only the residual delta is aligned across modalities. Fig.~\ref{fig:motivation} illustrates this shift.

We introduce \method, an anatomy-conditioned VLP framework. \method maintains a whole-study representation and mask-gated organ representations in parallel, grounds them with the whole report and with organ-specific report snippets, and uses a fixed pathology-to-organ map to keep representation learning, burden ordering, and evaluation anatomically consistent.
Burden-Direction Alignment (BDA) is designed to supply the missing direction signal. For each organ-mapped pathology, a fixed rule-based scorer mines confidence-filtered lower-to-higher relations from labels and report language across different patients, and each relation is applied separately at the whole-study scope and, when valid, at the corresponding organ scope. At its core, BDA aligns the pathology-conditioned image and report delta and favors the selected direction over its reverse. 
Two auxiliary constraints keep that orientation well posed: an endpoint consistency supplies an additional training signal, and a composition constraint keeps pairwise directions coherent across ordered triplets of increasing burden.
All BDA modules act in feature space and are discarded after training, leaving inference unchanged.

We evaluate \method on CT-RATE and RAD-ChestCT through zero-shot abnormality diagnosis, frozen-encoder linear probing, image–image and report–image retrieval, component and aligner ablations, and qualitative burden-order analysis. \method reaches 85.6 AUROC on CT-RATE, 7.8 points above the strongest prior method, and transfers to external RAD-ChestCT at 72.7 AUROC. It further raises frozen-encoder linear probing to 87.5 AUROC and report–image R@5 from 2.9 to 18.8.

Our contributions are:
\begin{itemize}
    \item We introduce \method, an anatomy-conditioned CT vision–language framework that learns not only which findings a report describes but how much disease is present, by mining weak ordered relations between different patients in place of the longitudinal data such order would otherwise require.

    \item We formulate Burden-Direction Alignment, which conditions both modalities on a pathology embedding, removes between-patient variation with a target-conditioned aligner, and aligns the resulting image and report delta under directional and composition constraints with auxiliary endpoint consistency.
    
    \item We show that \method reaches 85.6 AUROC on CT-RATE and 72.7 AUROC on external RAD-ChestCT, with consistent gains in frozen-encoder linear probing and retrieval, and we isolate the contribution of each constraint and of the aligner through controlled ablations.
\end{itemize}

\section{Related Work}
\label{sec:related_work}

\begin{figure*}[!t]
\centering
\includegraphics[width=\textwidth]{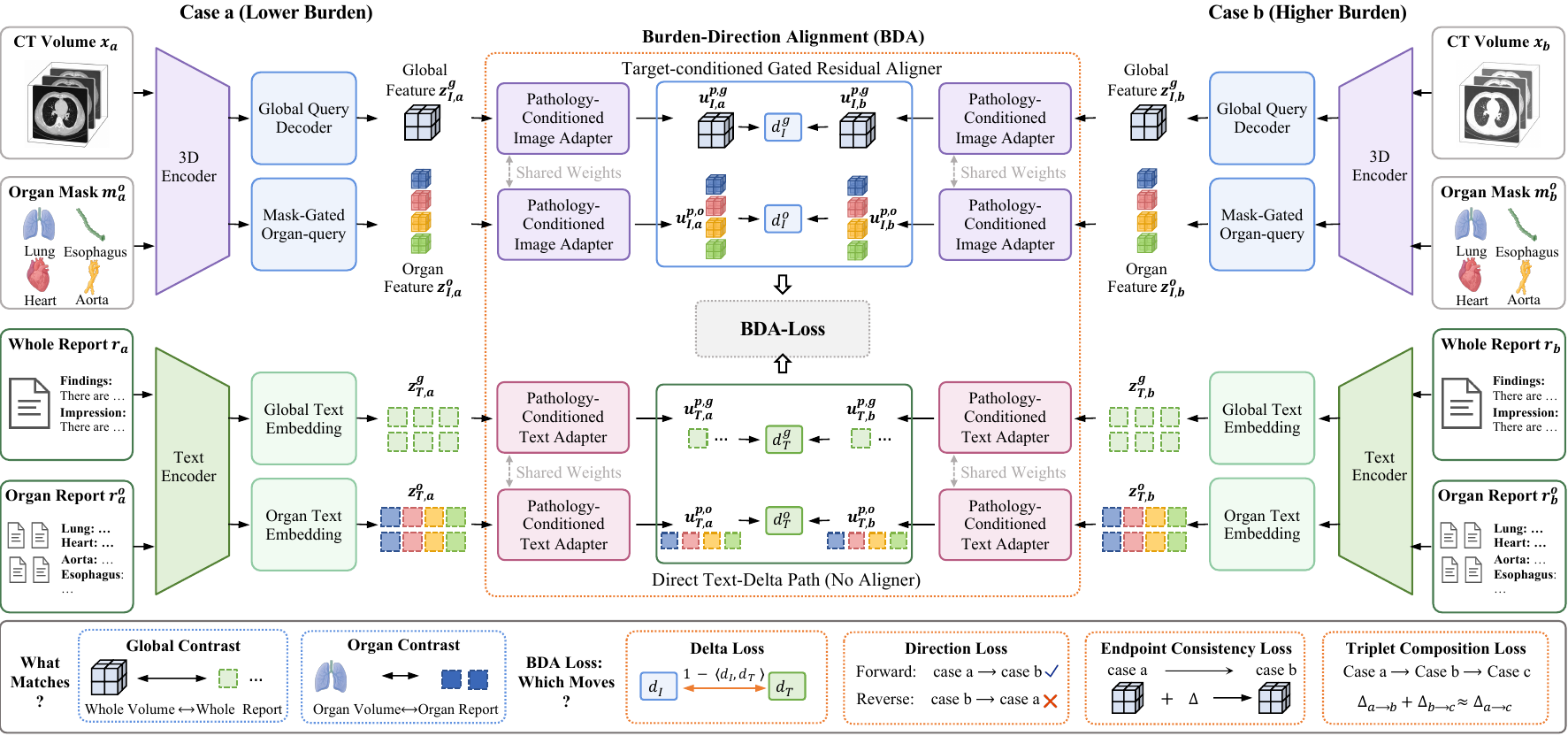}
\caption{
Architecture of \method. A mined relation orders two different patients, $a$ (lower burden) and $b$ (higher), for pathology $p$. Shared encoders embed both studies at whole-study and mask-gated organ scopes, where paired contrastive objectives fix correspondence (\emph{what matches}). Burden-Direction Alignment (BDA) adds order (\emph{which moves}), aligning image and report delta directions $d_I$ and $d_T$ at each scope. BDA is training-only.
}
\label{fig:architecture}
\end{figure*}

\subsection{Medical VLP as correspondence learning.}
Contrastive VLP learns transferable representations from paired or semantically matched image--text data \citep{radford2021learning,zhai2022lit,zhang2020contrastive,wang2022medclip,tiu2022chexzero,gong2021crossmodal,gong2022vqamix}. Medical models exploit report semantics through radiology-specific text modeling, global/local or multi-granular alignment, and structured entity/knowledge supervision \citep{boecking2022making,huang2021gloria,wang2022mgca,muller2022lovt,wu2023medklip,jain2021radgraph}. Broader biomedical models scale this paradigm to millions of scientific image--text pairs \citep{zhang2023biomedclip,lin2023pmcclip}. Despite these advances, existing VLP objectives primarily organize \emph{what matches}, rather than explicitly constraining a consistent lower-to-higher burden direction among cases sharing the same pathology.

\subsection{Volumetric and anatomy-aware CT VLP.}
Volumetric CT methods establish whole-volume--report alignment, transfer radiographic knowledge, or incorporate structured, generative, and image-only objectives for representation learning
\citep{hamamci2024generalist,bai2024m3d,cao2024biud,langlotz2024merlin,lai2025brgsa,wald2025colipri}.
CT-GLIP and fVLM localize correspondence by constructing organ-level image--text pairs or extracting mask-guided organ representations
\citep{lin2024ctglip,shui2025fvlm}. These methods expose anatomy-specific evidence.
Rather than introducing another correspondence granularity, \method uses the whole-study and pathology-mapped organ scopes to learn weak cross-patient burden direction by aligning ordered image--text deltas.

\subsection{Relative order and temporal supervision.}
Pairwise ranking and severity-aware representation learning impose order from preferences or explicit severity annotations
\citep{burges2005learning,cong2024conpro}.
Longitudinal medical VLP derives change supervision from patient-matched prior/current radiographs and their reports
\citep{bannur2023learning,yang2025tempavlp}.
These methods obtain order from explicit severity supervision.
In contrast, BDA mines confidence-filtered lower-to-higher pathology-burden relations from labels and report language across cross-sectional CT studies from different patients.

\section{Method}
\label{sec:method}

\subsection{Problem Formulation and Overview}

Each study $i$ contains a CT volume $x_i$, organ masks $m_i=\{m_i^o\}_{o\in\mathcal{O}}$, a whole report $r_i$, organ-specific report snippets $\{r_i^o\}_{o\in\mathcal{O}}$, and CT-RATE abnormality labels $y_i\in\{0,1\}^{18}$. We use four organ groups, $\mathcal{O}=\{\text{lung},\text{heart},\text{esophagus},\text{aorta}\}$, together with a global scope $g$. Here, a \emph{scope} $q\in\{g\}\cup\mathcal{O}$ denotes paired image and text representations at one anatomical level. A fixed pathology-to-organ map $\sigma$ assigns 16 of the 18 labels to one organ: 11 pulmonary findings to the lung, 3 cardiac findings to the heart, hiatal hernia to the esophagus, and arterial wall calcification to the aorta. BDA is applied to these 16 organ-mapped pathologies, while medical material and lymphadenopathy remain global-only. A selected relation $(a,b,p)$ encodes a lower-to-higher order between two different patients for pathology $p$.
Fig.~\ref{fig:architecture} summarizes the framework. \method organizes each study into complementary anatomy-conditioned scopes. A whole-study representation retains global context, while four mask-gated organ queries produce organ-specific representations; paired image--text objectives align each visual representation with its whole-report or organ-snippet counterpart. These scopes define \emph{where} correspondence is learned; by aligning pathology-conditioned image and text deltas, BDA defines \emph{how} lower-to-higher pathology-burden order is imposed at the global and organ scopes.

\subsection{Anatomy-Conditioned Scopes}

\paragraph{Global and organ-specific representations.}
A 3D encoder produces the feature map $F_i=E_\theta(x_i)$. A learned global query decoder $G$ aggregates the feature map into the whole-study feature $g_i=G(F_i)$. For each organ $o$, the mask $m_i^o$ is downsampled to the feature-map resolution and used as an attention mask; the mask-gated organ-query aggregator $H$, with one learned query per organ, pools the valid anatomical features into $h_i^o=H(F_i,m_i^o)$. A shared text encoder maps the whole report and organ-specific report snippets to $t_i^g=E_\phi(r_i)$ and $t_i^o=E_\phi(r_i^o)$, respectively. A global visual projection head $P_I^g$, an organ-scope visual projection head $P_I^o$, and a shared text projection head $P_T$ map these features to the embeddings used by the correspondence objectives:
\begin{equation}
\begin{aligned}
z_{I,i}^g&=P_I^g(g_i),&
z_{T,i}^g&=P_T(t_i^g),\\
z_{I,i}^o&=P_I^o(h_i^o),&
z_{T,i}^o&=P_T(t_i^o).
\end{aligned}
\end{equation}
Invalid organ masks are excluded by a validity indicator $v_i^o$. The organ visual projection $P_I^o$ is shared across organs, while $P_T$ is shared by whole reports and organ snippets.

\paragraph{Global and organ correspondence.}
Let $Z_I^g$ and $Z_T^g$ stack a mini-batch of projected whole-study representations. Global correspondence is learned with a symmetric image--text contrastive loss using diagonal image--report pairs:
\begin{equation}
\small
\mathcal{L}_{\mathrm{global}}=
\tfrac{1}{2}\!\left[
\operatorname{CE}(Z_I^g{Z_T^g}^{\!\top},I)+
\operatorname{CE}(Z_T^g{Z_I^g}^{\!\top},I)
\right].
\end{equation}
For each organ $o$, let $\mathcal{B}_o=\{i\in\mathcal{B}:v_i^o=1\}$ denote the valid studies, and let $Z_I^o$ and $Z_T^o$ stack their organ image and snippet representations. We define the organ-level similarity matrix as $S^o=Z_I^o{Z_T^o}^{\!\top}/\tau_o$. Organ correspondence is learned with the same symmetric paired contrastive objective:
\begin{equation}
\small
\mathcal{L}_{\mathrm{organ}}
=
\frac{1}{2|\mathcal{O}_{\mathcal{B}}|}
\sum_{o\in\mathcal{O}_{\mathcal{B}}}
\left[
\operatorname{CE}(S^o,I_o)
+
\operatorname{CE}\!\left((S^o)^{\!\top},I_o\right)
\right],
\end{equation}
where $I_o$ is the diagonal pairing target over $\mathcal{B}_o$, and $\mathcal{O}_{\mathcal{B}}$ contains organs with sufficient valid image--snippet pairs in the mini-batch. Together, $\mathcal{L}_{\mathrm{global}}$ and $\mathcal{L}_{\mathrm{organ}}$ establish \emph{what matches} at the whole-study and organ scopes, while cross-patient lower-to-higher order is introduced only by BDA. 

\subsection{Burden-Direction Alignment}

\paragraph{Rule-based burden scoring and cross-patient relation mining.}
For each organ-mapped pathology $p$, a fixed, non-learned scorer $\phi_p$ combines the label $y_{i,p}$ with the whole report and corresponding organ-specific report snippet to produce a burden-evidence score $s_i^p$ and confidence $\rho_i^p$. Both \(s_i^p\in\mathbb{R}_{\ge0}\) and \(\rho_i^p\in\mathbb{R}_{\ge0}\) are additive, non-normalized scores rather than probabilities. Different patients form an eligible candidate relation when
\begin{equation}
\small
\begin{aligned}
(s_i^p,\rho_i^p)&=\phi_p(y_{i,p},r_i,r_i^{\sigma(p)}),\\
(a,b,p)\in\widetilde{\mathcal{P}}
&\Longleftrightarrow
\begin{gathered}
a\neq b,\quad s_b^p-s_a^p\geq\delta,\\
s_a^p,s_b^p\geq s_{\min},\quad
\rho_a^p,\rho_b^p\geq\eta.
\end{gathered}
\end{aligned}
\end{equation}
Here, $\delta$ requires sufficient separation between the two burden-evidence scores, $s_{\min}$ requires both endpoints to contain adequate evidence for pathology $p$, and $\eta$ filters low-confidence relations. The scorer parses pathology mentions, negation patterns, burden or severity descriptors, size or comparative cues, and uncertainty language. Its score supports relative ordering.
Within mini-batches, eligible candidates from all active pathologies are assigned the reliability weight $w_{ab}^p=(s_b^p-s_a^p)\times\min(\rho_a^p,\rho_b^p)$, and the top-$K$ candidates form the selected relation set $\mathcal{P}\subseteq\widetilde{\mathcal{P}}$. Each selected triple defines a cross-patient relation $a\!\rightarrow\!b$, in which study $a$ carries weaker burden cues than study $b$ for pathology $p$.

Relation mining is scope-independent: each triple $(a,b,p)$ is selected once and then evaluated at the whole-study scope $q=g$ and the corresponding organ scope $q=\sigma(p)$. At $q=g$, the pathology embedding $e_p$ conditions the two whole-study representations, yielding a pathology-indexed global burden direction rather than a generic whole-study burden axis. At $q=\sigma(p)$, the same relation constrains the corresponding organ-specific representations, grounding the burden direction in the mapped anatomy.
\paragraph{Pathology-conditioned delta construction.}
For each selected relation $(a,b,p)$ and scope $q\in\{g,\sigma(p)\}$, we first condition the corresponding image and text representations on a learned pathology embedding $e_p$ using modality-specific gated residual adapters:
\begin{equation}
\small
u_{M,i}^{p,q}=C_M(z_{M,i}^{q};e_p),\qquad M\in\{I,T\},\quad i\in\{a,b\}.
\end{equation}
For each modality, $C_M$ concatenates the selected representation with $e_p$; modality-specific two-layer update and sigmoid-gate networks produce a gated residual that is added to the input. The image and text adapters are shared across studies, anatomical scopes, and pathologies, with $p$ changing only the embedding $e_p$ rather than selecting an independent network. Each relation is therefore conditioned within one active whole-study or organ-local scope.

\begin{figure}[!t]
\centering
\includegraphics[width=\linewidth]{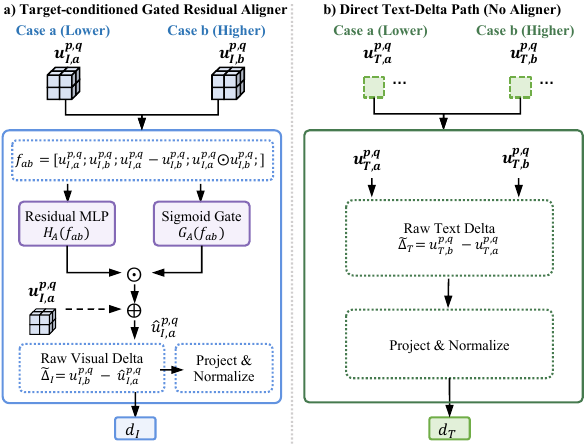}
\caption{
Delta construction in BDA for pathology $p$ at scope $q$. A target-conditioned aligner maps endpoint $a$ (lower burden) toward $b$ (higher) before differencing, giving $d_I$, while the aligner-free text path gives $d_T$. The endpoints are different patients, so only the residual difference isolates burden from habitus and protocol.
}
\label{fig:vpa_delta_paths}
\end{figure}

Before measuring visual difference, a target-conditioned aligner produces a target-aware visual source, as shown in Fig.~\ref{fig:vpa_delta_paths}(a). The aligned visual source, pre-projection image and text differences, and normalized projected directions $d_M$ for $M\in\{I,T\}$ are
\begin{equation}
\small
\label{eq:pathology_delta}
{\setlength{\jot}{2pt}
\begin{alignedat}{2}
\widehat{u}_{I,a}^{p,q}
&=A(u_{I,a}^{p,q},u_{I,b}^{p,q}), &\;
\widetilde{\Delta}_I^{a\to b,p,q}
&=u_{I,b}^{p,q}-\widehat{u}_{I,a}^{p,q},\\
\widetilde{\Delta}_T^{a\to b,p,q}
&=u_{T,b}^{p,q}-u_{T,a}^{p,q}, &\;
d_M^{a\to b,p,q}
&=\operatorname{norm}\!\bigl(\widetilde{\Delta}_M^{a\to b,p,q}\bigr).
\end{alignedat}}
\end{equation}
The aligner $A$ concatenates the visual source, target, their signed difference, and element-wise interaction; a two-layer MLP predicts a residual update modulated by a learned sigmoid gate and added to the source. For the forward relation $a\!\rightarrow\!b$, only the source feature is transformed, while the target feature remains unchanged. The text branch bypasses the aligner and uses the direct difference between pathology-conditioned report representations, retaining the textual difference as a direct semantic reference. The modality-specific projectors map both differences into a shared delta space, and $\operatorname{norm}$ denotes $\ell_2$ normalization.

\paragraph{Direction-aware and geometric constraints.}
The deltas above define pathology-conditioned cross-patient difference. The core BDA term makes this difference cross-modal by aligning the unit image and text directions:
\begin{equation}
\small
\mathcal{L}_{\Delta}
=
1-\left\langle
d_I^{a\rightarrow b,p,q},
d_T^{a\rightarrow b,p,q}
\right\rangle.
\end{equation}
Because both directions are $\ell_2$-normalized, $\mathcal{L}_{\Delta}$ performs cosine direction matching rather than batch-level contrastive learning; $\mathcal{L}_{\mathrm{global}}$ and $\mathcal{L}_{\mathrm{organ}}$ preserve the absolute image--text correspondence.
We omit the common indices $(p,q)$ in the following auxiliary terms. To distinguish the selected orientation from undirected similarity, let $s_{\mathrm{fwd}}=\langle d_I^{a\rightarrow b},d_T^{a\rightarrow b}\rangle$ and $s_{\mathrm{rev}}=\langle d_I^{b\rightarrow a},d_T^{a\rightarrow b}\rangle$. Here, $d_I^{b\rightarrow a}$ is recomputed by swapping the visual source and target through the same aligner and delta projector, rather than being defined as $-d_I^{a\rightarrow b}$. The auxiliary constraints are
\begin{equation}
\small
\begin{aligned}
\mathcal{L}_{\mathrm{dir}}
&=
\left[\mu-\left(s_{\mathrm{fwd}}-s_{\mathrm{rev}}\right)\right]_+,\\
\mathcal{L}_{\mathrm{end}}
&=
\operatorname{MSE}\!\left(
\Psi\!\left([\widehat{u}_{I,a};\widetilde{\Delta}_I^{a\rightarrow b}]\right),
u_{I,b}
\right),\\
\mathcal{L}_{\mathrm{comp}}
&=
\operatorname{MSE}\!\left(
d_I^{a\rightarrow c},
\operatorname{norm}\!\left(
d_I^{a\rightarrow b}+d_I^{b\rightarrow c}
\right)
\right).
\end{aligned}
\end{equation}
Here, $[x]_+=\max(0,x)$. The direction term enforces $s_{\mathrm{fwd}}\geq s_{\mathrm{rev}}+\mu$ while keeping the textual reference fixed in the selected $a\!\rightarrow\!b$ direction. The endpoint head $\Psi$ maps $[\widehat{u}_{I,a};\widetilde{\Delta}_I^{a\rightarrow b}]\in\mathbb{R}^{2D}$ to the observed higher-burden visual endpoint, so $\mathcal{L}_{\mathrm{end}}$ serves as an auxiliary latent endpoint-consistency regularizer rather than an image-reconstruction. For an eligible same-pathology triplet with $s_a^p<s_b^p<s_c^p$, $\mathcal{L}_{\mathrm{comp}}$ encourages the direct $a\!\rightarrow\!c$ direction to agree with the normalized two-step path $a\!\rightarrow\!b\!\rightarrow\!c$. 

Using the reliability weight $w_{ab}^p$ defined during relation mining, let $\mathcal{P}_q$ denote the selected relations whose representations are valid at scope $q$. The pairwise and scope-level objectives are
\begin{equation}
\small
\begin{aligned}
\ell_{ab}^{p,q}
&=
\mathcal{L}_{\Delta}
+\lambda_{\mathrm{dir}}\mathcal{L}_{\mathrm{dir}}
+\lambda_{\mathrm{end}}\mathcal{L}_{\mathrm{end}},\\
\mathcal{L}_{\mathrm{BDA}}^q
&=
\frac{
\sum_{(a,b,p)\in\mathcal{P}_q}
w_{ab}^p\ell_{ab}^{p,q}
}{
\sum_{(a,b,p)\in\mathcal{P}_q}
w_{ab}^p
}
+\lambda_{\mathrm{comp}}
\overline{\mathcal{L}}_{\mathrm{comp}}^q.
\end{aligned}
\end{equation}
$\overline{\mathcal{L}}_{\mathrm{comp}}^q$ averages eligible same-pathology triplets at scope $q$ and is zero when none exists; a scope with $\mathcal{P}_q=\varnothing$ is omitted from the batch objective. Thus, the same mined relation is supervised at the whole-study scope and, when valid, at the corresponding pathology-mapped organ scope.

\subsection{Training Objective}

Before cross-modal training, we warm up the visual branch with $\mathcal{L}_{\mathrm{warm}}=\mathcal{L}_{\mathrm{anchor}}+\lambda_{\mathrm{rank}}\mathcal{L}_{\mathrm{rank}}$. $\mathcal{L}_{\mathrm{anchor}}$ applies label-supervised prediction at the whole-study scope and at valid pathology-mapped organ scopes. At both scopes, $\mathcal{L}_{\mathrm{rank}}$ uses a margin hinge so that a label-positive study scores above a label-negative study for the same pathology. This stage establishes coarse disease-sensitive representations by separating pathology presence from absence, but does not order different burden states among studies sharing the same pathology; the lower-to-higher relations used by BDA are subsequently mined from labels and report cues. During image--text training, the 3D backbone is frozen, while the global decoder, organ-query module, projection layers, text encoder, and BDA-specific modules remain trainable. The training objective combines paired whole-study and organ-local correspondence with cross-patient burden order:
\begin{equation}
\small
\begin{aligned}
\mathcal{L}_{\mathrm{train}}={}&
\underbrace{\mathcal{L}_{\mathrm{global}}+
\lambda_{\mathrm{organ}}\mathcal{L}_{\mathrm{organ}}}
_{\text{correspondence}}\\
&+
\underbrace{\lambda_{\mathrm{BDA}}
\frac{1}{|\mathcal{Q}_{\mathcal{B}}|}
\sum_{q\in\mathcal{Q}_{\mathcal{B}}}
\mathcal{L}_{\mathrm{BDA}}^q}
_{\text{cross-patient burden order}} .
\end{aligned}
\end{equation}
Here, $\mathcal{Q}_{\mathcal{B}}=\{q:\mathcal{P}_q\neq\varnothing\}$ contains the whole-study and pathology-mapped organ scopes with at least one selected relation valid in the current mini-batch. When $\mathcal{Q}_{\mathcal{B}}$ is empty, the aggregate BDA term is set to zero.
\begin{table*}[!t]
\centering
\small
\setlength{\tabcolsep}{12pt}
\begin{tabular}{l|cccc|cccc}
\toprule
 & \multicolumn{4}{c|}{CT-RATE$\uparrow$} &
\multicolumn{4}{c}{RAD-ChestCT$\uparrow$} \\
\cline{2-9}
Method & \auc{} & ACC & F1 & Prec. & \auc{} & ACC & F1 & Prec. \\
\midrule
CT-Net~\citep{draelos2021radchestct} & 60.3 & 58.1 & 63.1 & 23.9 & 54.4 & 54.0 & 58.7 & 28.5 \\
CT-CLIP~\citep{hamamci2024generalist} & 73.1 & 66.8 & 70.7 & 32.3 & 62.9 & 59.5 & 64.2 & 33.6 \\
BIUD~\citep{cao2024biud} & 71.3 & 68.1 & 71.6 & 33.8 & 62.9 & 60.6 & 65.2 & 33.7 \\
Merlin~\citep{langlotz2024merlin} & 72.8 & 67.2 & 70.9 & 33.7 & 64.4 & 61.9 & 66.3 & 34.8 \\
COLIPRI-C$^\dagger$~\citep{wald2025colipri} & 76.3 & -- & -- & -- & 69.1 & -- & -- & -- \\
fVLM~\citep{shui2025fvlm} & 77.8 & 71.8 & 75.1 & 37.9 & 68.0 & 64.7 & 68.8 & 37.4 \\
\midrule
\rowcolor[gray]{.92}\textbf{\method (Ours)} & \textbf{85.6} & \textbf{79.0} & \textbf{81.2} & \textbf{46.2} & \textbf{72.7} & \textbf{67.7} & \textbf{71.9} & \textbf{39.7} \\
\bottomrule
\end{tabular}
\caption{
Zero-shot abnormality diagnosis. No downstream classifier or evaluation-set tuning is used. \auc{}, ACC (accuracy), F1 and Prec.\ (precision), in \%; ``--'': not reported; $^\dagger$: contrastive-only variant. Higher is better.
}
\label{tab:diagnosis_results}
\end{table*}

\begin{table}[!t]
\centering
\small
\setlength{\tabcolsep}{2.5pt}
\begin{tabular}{l|cccc}
\toprule
& \multicolumn{2}{c}{CT-RATE}
& \multicolumn{2}{c}{RAD-ChestCT} \\
\cmidrule(lr){2-3}\cmidrule(l){4-5}
Method
& \small AUPRC & \small AUROC
& \small  AUPRC & \small AUROC \\
\midrule
Curia~\citep{dancette2025curia}
& 45.9 & 78.0 & 39.2 & 65.7 \\

CT-CLIP
& -- & 75.1 & -- & 64.7 \\

CT-FM~\citep{ct-fm}
& 53.5 & 82.1 & 42.4 & 68.5 \\

Merlin
& 54.8 & 82.6 & 45.3 & 70.9 \\

COLIPRI-C
& 55.1 & 82.6 & 46.3 & 71.2 \\
\midrule
\rowcolor[gray]{.92}\textbf{\method (Ours)} & \textbf{64.7} & \textbf{87.5}& \textbf{52.2} & \textbf{72.5} \\
\bottomrule
\end{tabular}
\caption{
Frozen-encoder linear probing. A linear classifier is fitted on frozen image features. AUPRC and \auc{} in \%; ``--'': not reported. Higher is better. The advantage of \method persists with the encoder frozen.
}
\label{tab:linear_probe}
\end{table}

\section{Experiments}
\label{sec:experiments}

\subsection{Datasets and Metrics}

\paragraph{Datasets.}
We evaluate \method on two public datasets: CT-RATE and RAD-ChestCT. CT-RATE contains 25,692 non-contrast 3D scans from 21,304 patients, expanded to 50,188 reconstructed volumes with paired radiology reports and abnormality labels \citep{hamamci2024generalist}. RAD-ChestCT is a non-contrast chest CT abnormality dataset from Duke University, comprising 36,316 volumes collected between 2012 and 2017 with 83 abnormality labels \citep{draelos2021radchestct}. Its public subset contains 3,630 scans.

\paragraph{Metrics.}
We consider four evaluation tasks: zero-shot abnormality diagnosis, frozen-encoder linear probing, image--image retrieval, and report--image retrieval. For zero-shot diagnosis evaluation, we follow the protocol of CT-CLIP~\cite{hamamci2024generalist}. These tasks assess zero-shot diagnostic prediction, linear separability, and retrieval-based representation quality. For zero-shot diagnosis, we report AUROC, ACC, F1, and precision; for linear probing, we report AUPRC and AUROC; for image--image retrieval, we report MAP@5/10/50; and for report--image retrieval, we report Recall@5/10/50/100. Reported baseline results follow their published protocols, and missing metrics are marked as ``--''. We additionally inspect cross-patient burden order through one pathology-specific three-study chain.

\subsection{Implementation Details}

The CT encoder is a 3D ResNet-18. Organ masks are grouped into lung, heart, esophagus, and aorta and used to select organ-specific evidence from the 3D feature map. A four-query global decoder and four mask-gated organ queries produce 512-dimensional visual representations, while CXR-BERT encodes whole reports and organ-specific report snippets as 768-dimensional text features. Across visual warm-up and image--text training, we set $\lambda_{\mathrm{rank}}=\lambda_{\mathrm{organ}}=0.10$, $\lambda_{\mathrm{BDA}}=0.64$, $\lambda_{\mathrm{dir}}=0.50$, $\lambda_{\mathrm{end}}=0.25$, $\lambda_{\mathrm{comp}}=0.03$, and  \(\tau_o=0.07\). During image--text training, we use AdamW with batch size 4 and learning rates of $10^{-5}$ and $10^{-4}$ for the text encoder and projection modules, respectively. BDA uses $\delta=\eta=s_{\min}=0.75$, $K=32$, at most eight eligible triplets per mini-batch, a 256-dimensional delta space, and margin $\mu=0.10$. BDA losses are averaged equally across active whole-study and organ scopes. All experiments were conducted using NVIDIA A100 GPUs.

\subsection{Main Results}
\paragraph{Zero-shot abnormality diagnosis.} Table~\ref{tab:diagnosis_results} compares \method with existing CT methods on CT-RATE and RAD-ChestCT. Here, zero-shot follows the CT VLP inference protocol: no downstream classifier is fitted and no evaluation-set fine-tuning is performed. On CT-RATE, \method achieves the highest result among the compared methods for every metric, reaching 85.6 \auc{}, 79.0 ACC, 81.2 F1, and 46.2 precision. Relative to fVLM, these scores improve by 7.8, 7.2, 6.1, and 8.3 points; \method also surpasses the contrastive-only COLIPRI-C variant by 9.3 \auc{} points. On RAD-ChestCT, \method obtains 72.7 \auc{}, 67.7 ACC, 71.9 F1, and 39.7 precision, exceeding the prior \auc{} in the table by 3.6 points and outperforming fVLM by 4.7 \auc{} points. The gains across both datasets demonstrate stronger in-domain diagnosis and external zero-shot transfer.

\paragraph{Frozen-encoder linear probing.} Table~\ref{tab:linear_probe} evaluates the discriminative quality of the frozen image representation under a linear classifier. On CT-RATE, \method achieves 64.7 AUPRC and 87.5 AUROC, surpassing the strongest baseline by 9.6 and 4.9 points, respectively. On RAD-ChestCT, it reaches 52.2 AUPRC and 72.5 AUROC, improving over the strongest baseline by 5.9 and 1.3 points. These results show that \method produces more linearly separable features for both in-domain and external abnormality classification.

\begin{table}[!t]
\centering
\small
\setlength{\tabcolsep}{2pt}
\begin{tabular}{p{0.28\linewidth}|p{0.30\linewidth}p{0.34\linewidth}}
\toprule
Method &
\shortstack{Img2Img \\ MAP@5/10/50$\uparrow$} &
\shortstack{Rpt2Img \\Recall@5/10/50/100$\uparrow$} \\
\midrule
CT-Net & 59.4 / 48.1 / 40.7 & -- / -- / -- / -- \\
Merlin & 62.6 / 51.3 / 43.9 & 1.5 / 2.7 / 7.7 / 12.7 \\
CT-CLIP & 68.3 / 57.2 / 48.9 & 2.9 / 5.0 / 18.0 / 28.7 \\
\ -- ClassFine & 67.9 / 56.8 / 48.5 & -- / -- / -- / -- \\
fVLM & 49.1 / 36.8 / 26.0 & 0.8 / 1.5 / 4.9 / 8.3 \\
\midrule
\rowcolor[gray]{.92}\textbf{\method (Ours)} & \textbf{71.2 / 61.6 / 54.2} & \textbf{18.8 / 26.3 / 51.8 / 64.3} \\
\bottomrule
\end{tabular}
\caption{CT-RATE retrieval. Image--image MAP measures abnormality-overlap retrieval, while report--image Recall evaluates paired cross-modal retrieval.}
\label{tab:retrieval_results}
\end{table}

\paragraph{Image--image retrieval.} Table~\ref{tab:retrieval_results} shows that \method achieves mAP@5/10/50 of 71.2/61.6/54.2. Compared with CT-CLIP, these results improve by 2.9, 4.4, and 5.3 points, respectively. The increasing margin at deeper retrieval cutoffs demonstrates that the learned image space preserves abnormality-aware neighborhood structure beyond the nearest scans. Together with the diagnosis and linear-probing results, this confirms that anatomy-conditioned burden-order learning improves both category discrimination and the organization of CT representations.

\paragraph{Report--image retrieval.} For report--image retrieval, \method achieves R@5/10/50/100 of 18.8/26.3/51.8/64.3, compared with 2.9/5.0/18.0/28.7 for CT-CLIP. These correspond to gains of 15.9, 21.3, 33.8, and 35.6 points, respectively. The substantial improvements across retrieval depths demonstrate that \method preserves strong report--scan correspondence while learning anatomy-conditioned pathology-burden directions. BDA therefore complements the global and organ-local correspondence objectives and strengthens the resulting cross-modal representation space.

\subsection{Ablation and Analysis}

\paragraph{BDA loss ablation.} Table~\ref{tab:vpa_components} compares the correspondence-only baseline with the cumulative addition of the four BDA losses under the same representation architecture and optimization protocol. Without BDA supervision, the model obtains 76.7 \auc{}. Adding the core image--text delta-alignment loss $\mathcal{L}_{\Delta}$ raises performance to 83.9, a gain of 7.2 points. Forward--reverse discrimination $\mathcal{L}_{\mathrm{dir}}$ and endpoint consistency $\mathcal{L}_{\mathrm{end}}$ further improve it to 84.3 and 84.5, respectively. The complete objective reaches 85.6 \auc{}. In particular, the 1.1-point gain from $\mathcal{L}_{\mathrm{comp}}$ shows that pairwise delta alignment benefits substantially from path consistency across ordered $a\!\rightarrow\!b\!\rightarrow\!c$ triplets.

\begin{table}[!t]
\centering
\small
\setlength{\tabcolsep}{12.6pt}
\begin{tabular}{cccc|c}
\toprule
$L_{\Delta}$ & $L_{dir}$ & $L_{end}$ & $L_{comp}$ &
\auc{}$\uparrow$ \\
\midrule
-- & -- & -- & -- &
76.7 \\
\checkmark & -- & -- & -- &
83.9 \\
\checkmark & \checkmark & -- & -- &
84.3 \\
\checkmark & \checkmark & \checkmark & -- &
84.5 \\
\rowcolor[gray]{.92} \checkmark & \checkmark & \checkmark & \checkmark &
\textbf{85.6} \\
\bottomrule
\end{tabular}
\caption{
BDA loss ablation on CT-RATE. Columns add delta alignment, forward--reverse direction, endpoint consistency and ordered-triplet composition in turn.
}

\label{tab:vpa_components}
\end{table}

\begin{figure}[!t]
\centering
\includegraphics[width=\linewidth]{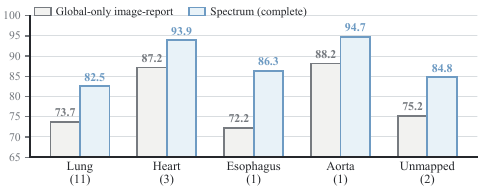}
\caption{
Global-prediction AUROC by organ group on CT-RATE. \method is compared with global-only image--report alignment, a matched internal control rather than a prior method. Groups follow the pathology-to-organ map, with pathologies per group in parentheses and ``unmapped'' the two global-only labels.
}
\label{fig:anatomy_evidence}
\end{figure}

\paragraph{Aligner controls.} Table~\ref{tab:aligner_controls} keeps the complete BDA objective fixed and changes only how the lower-burden image representation is aligned before computing its delta. Removing the higher-burden representation from the aligner input reduces \auc{} from 85.6 to 84.5, while bypassing the aligner entirely yields 84.2. Thus, target context contributes 1.1 points over the target-blind variant, and the complete target-conditioned aligner contributes 1.4 points over direct source-to-target subtraction.

\paragraph{Organ-group behavior.} Fig.~\ref{fig:anatomy_evidence} groups the 18 abnormalities predicted from the global representation by their pathology-to-organ assignment. Compared with global-only image--report alignment, the complete framework improves lung, heart, esophagus, aorta, and the two unmapped pathologies by 8.8, 6.7, 14.1, 6.5, and 9.6 \auc{} points, raising the overall score from 76.8 to 85.6. The positive gain in every group demonstrates that the improvement is not driven by a single dominant organ or pathology family. Notably, the two unmapped abnormalities also improve, showing that the anatomy-conditioned and burden-order training signals are encoded beyond their directly mapped organ scopes in the shared whole-study representation.

\begin{figure}[t]
\centering
\includegraphics[width=\linewidth]{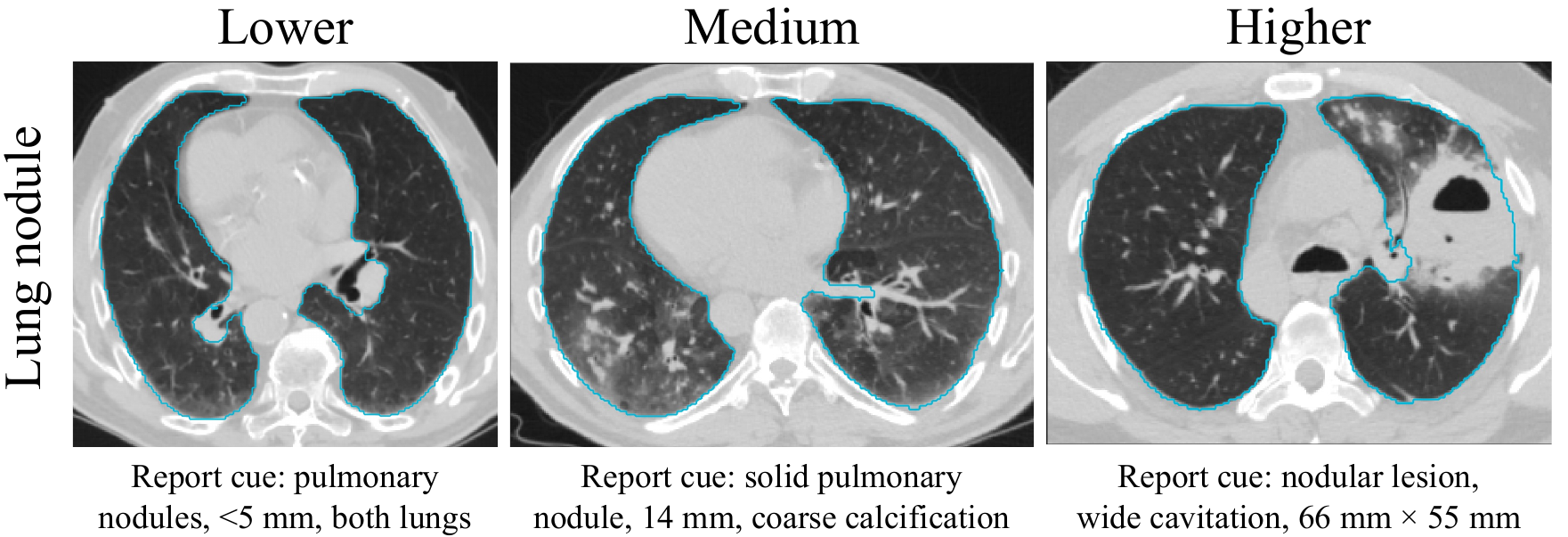}
\caption{
Model-supported burden order for lung nodule. BDA mines these three studies, one per patient, as a lower--medium--higher chain; the report size cues rise accordingly, and cyan contours delineate the lungs. All three forward--reverse similarity gaps are positive, so the trained model prefers the displayed direction to its reverse. The order is cross-patient, not longitudinal.
}
\label{fig:vpa_triplet_case}
\end{figure}

\paragraph{Model-derived burden-order geometry.}
For each ordered pair, we report the forward--reverse similarity difference $s_{\mathrm{fwd}}-s_{\mathrm{rev}}$, averaged over the whole-study and organ scopes. A positive value indicates that the visual transition agrees more strongly with the selected report direction than with its swapped reverse. For the lung-nodule triplet in Fig.~\ref{fig:vpa_triplet_case}, the lower-to-medium, medium-to-higher, and lower-to-higher gaps are $+0.034$, $+0.011$, and $+0.060$, respectively. A least-squares one-dimensional fit jointly summarizes the three pairwise gaps; within-triplet min--max normalization yields coordinates $0.000/0.716/1.000$, recovering the lower--medium--higher order.

\begin{table}[!t]
\centering
\small
\setlength{\tabcolsep}{8.0pt}
\begin{tabular}{l|cc|c}
\toprule
Aligner & Target input & Learned & \auc{}$\uparrow$ \\
\midrule
Identity & -- & -- & 84.2 \\
Target-blind & -- & \checkmark & 84.5 \\
\rowcolor[gray]{.92}Target-conditioned & \checkmark & \checkmark & \textbf{85.6} \\
\bottomrule
\end{tabular}
\caption{
Visual-aligner controls on CT-RATE. All rows use the complete BDA objective and differ only in how the lower-burden image representation is aligned before constructing its delta. Target input indicates whether the higher-burden representation is visible to the aligner; Identity bypasses the learned aligner while retaining the target representation for delta construction.
}
\label{tab:aligner_controls}
\end{table}

\FloatBarrier

\section{Discussion and Conclusion}
\label{sec:conclusion}

\method advances volumetric CT VLP from anatomy-aware correspondence to anatomy-conditioned burden direction. By mining confidence-filtered lower-to-higher relations from cross-sectional labels and reports, BDA aligns pathology-conditioned image and text deltas at the whole-study and pathology-mapped organ scopes, converting weak cross-patient cues into structured feature-space order. The complete framework raises CT-RATE \auc{} from 76.8 to 85.6 over matched global-only image--report alignment and achieves 72.7 \auc{} on external RAD-ChestCT; consistent gains in frozen-encoder linear probing, retrieval, and pathology-group analysis further demonstrate stronger and more transferable representations. These results establish weak cross-patient order as a scalable complement to anatomy-aware correspondence. More broadly, \method moves CT VLP beyond learning only \emph{what matches} toward encoding consistent lower-to-higher pathology-burden directions from routine cross-sectional data, without requiring longitudinal CT.

\bibliography{references}

\end{document}